\def\year{2022}\relax
\documentclass[letterpaper]{article} 
\usepackage{aaai22}  
\usepackage{times}  
\usepackage{helvet}  
\usepackage{courier}  
\usepackage[hyphens]{url}  
\usepackage{graphicx} 
\usepackage{natbib}  
\usepackage{caption} 
\DeclareCaptionStyle{ruled}{labelfont=normalfont,labelsep=colon,strut=off} 
\usepackage{algorithm}
\usepackage{mathtools}
\usepackage{amssymb}
\usepackage{multirow}
\usepackage{color}
\usepackage{subcaption}
\usepackage[title]{appendix}
\usepackage{algpseudocode}

\usepackage{newfloat}
\usepackage{listings}
\floatstyle{ruled}
\newfloat{listing}{tb}{lst}{}
\floatname{listing}{Listing}
\def\model{LeSICiN}

\algnewcommand\algorithmicforeach{\textbf{for each}}
\algdef{S}[FOR]{For}[1]{\algorithmicforeach\ #1\ \algorithmicdo}

\title{LeSICiN: A Heterogeneous Graph-based Approach for Automatic \\Legal Statute Identification from Indian Legal Documents}

\author{
    Shounak Paul,
    Pawan Goyal, 
    Saptarshi Ghosh 
}

\affiliations{
    Department of Computer Science and Engineering\\
    Indian Institute of Technology, Kharagpur, India\\
    shounakpaul95@kgpian.iitkgp.ac.in, pawang@cse.iitkgp.ac.in, saptarshi@cse.iitkgp.ac.in
}

\usepackage{bibentry}

\begin{document}

\maketitle

\begin{abstract}
The task of Legal Statute Identification (LSI) aims to identify the legal statutes that are relevant to a given description of Facts or evidence of a legal case. 
Existing methods only utilize the textual content of Facts and legal articles to guide such a task. However, the citation network among case documents and legal statutes is a rich source of additional information, which is not considered by existing models. 
In this work, we take the first step towards utilising both the text and the legal citation network for the LSI task.
We curate a large novel dataset for this task, including Facts of cases from several major Indian Courts of Law, and statutes from the Indian Penal Code (IPC). 
Modeling the statutes and training documents as a heterogeneous graph, our proposed model {\bf LeSICiN} can learn rich textual and graphical features, and can also tune itself to correlate these features. 
Thereafter, the model can be used to inductively predict links between test documents (new nodes whose graphical features are not available to the model) and statutes (existing nodes). 
Extensive experiments on the dataset show that our model comfortably outperforms several state-of-the-art baselines, by exploiting the graphical structure along with textual features. 
The dataset and our codes are available at \url{https://github.com/Law-AI/LeSICiN}.
\end{abstract}

\section{Introduction} \label{sec:intro}
The task of Legal Statute Identification (LSI) is important in the judicial field,
and involves identifying the possible set of {\it statutory laws} that are relevant, or might have been violated, given the {\it natural language description of the Facts of a situation}. 
This task needs to be performed at different stages of litigation by different experts, including police personnel, lawyers and judges. 
An automated system for LSI can greatly promote access to Law by the common masses.

\vspace{1mm}
\noindent {\bf Existing LSI methods and their limitations:}
Early methods for LSI modeled the task using statistical or simple machine learning methods~\cite{kort1957predicting,lauderdale2012supreme}. 
Recently, several neural architectures have modeled the problem as a text classification task, attempting to extract increasingly richer features from the text~\cite{luo2017learning,wang2018modeling,wang2019hierarchical,xu2020distinguish}.
Some of these methods simplify the task by trying to identify only the most relevant statute, no longer retaining its multi-label nature.

Historically maintained court case data have been popularly used to create datasets to train the LSI models~\cite{chalkidis2019neural,xiao2018cail2018}. However, almost all existing methods for this task rely only on the text of the Facts (and statutes) to perform the classification task. 
The existing methods do {\it not utilize the legal statute citation network} between court case documents and statutes,
that has been shown to be a rich source of legal knowledge that is useful for tasks such as estimating legal document similarity~\cite{hier-spc-net}. 
However, there has not been any attempt to utilize the legal citation network for the LSI task.

\vspace{1mm}
\noindent {\bf Graph formulation of LSI:} This work is the first attempt towards LSI by utilizing both the textual content of statutes/Facts and a {\it heterogeneous statute citation network}.
In this network, statutes and document are nodes of different types, and 
citation links exist between these nodes {\it iff} a particular Section is cited by a particular document. 
Additionally, statute nodes may also be connected via defined hierarchies specified in the written law itself. 
Given such a network, the task of LSI boils down to predicting the existence of links between statute-nodes and {\it newly entering} document-nodes.


However, most existing supervised network-based training methods for link prediction {\it do not work well for out-of-sample nodes}, or nodes that the model has not seen during training. 
Indeed, in our case, we can only provide the document nodes and their citation links during training time. 
While testing, every document-node is previously unseen to the model, and the model must predict whether links exist between a new document-node and the statute nodes. Thus, approaches that seek to learn effective node representations for link prediction will fail during test time in our setting. 
Rather, one has to capitalize on the Fact that two kinds of information are available at all times --- the texts corresponding to each statute/document, and the statute nodes themselves (and hierarchical links between them).

We, therefore, formulate the LSI problem as an {\bf inductive link prediction} task~\cite{hao2020inductive}  on the heterogeneous citation network, between the previously seen Section nodes and a new, unknown Fact node. 
We use a hybrid learning mechanism to learn two kinds of representations for each node -- attribute (generated from text) and structural (generated from network). By forcing the model to learn attribute representations that closely match structural representations of the same node, we ensure that the model can generalize better by generating more robust, feature-rich attribute representations for unseen document nodes during test time.

\vspace{1mm}
\noindent {\bf A new LSI dataset:}
There do not exist many good quality datasets for LSI in English. 
The ECHR dataset~\cite{chalkidis2019neural}  is considerably small ($\sim 11.5K$ documents/Facts) and does not provide the text of the statutes. 
Another popular dataset, CAIL~\cite{xiao2018cail2018} is much larger; it is, however, in the Chinese language.
Importantly, the {\it average no. of statutes cited per document} is quite low in both these datasets ($0.71$ for ECHR and $1.09$ for CAIL), i.e., most documents cite at most one statute (or none in the case of ECHR). 
However, the LSI problem is inherently multi-label, and these datasets do not reflect the true multi-label nature of the LSI problem.

We construct a new, fairly large dataset ($\sim 66K$ docs) from case documents and statutes (in English) from the Indian judiciary. 
This dataset, which we call the \textit{Indian Legal Statute Identification} (ILSI) dataset, 
captures the multi-label nature of the LSI problem more truly (details in Section~\ref{sec:expt}).

\vspace{1mm}
\noindent {\bf Contributions of present work:}
To sum up, our contributions are as follows:
(1)~To our knowledge, we are the first to use the statute citation network in conjunction with textual descriptions for the task of Legal Statute Identification. 
We propose a novel architecture \textbf{\model~(Legal Statute Identification using Citation Network)} for the task, that out-performs several state-of-the-art baselines for LSI (improvement of $19.2\%$ over the closest competitor). 
(2)~We construct a large-scale LSI dataset from Indian court case documents, where the task is to identify Sections of the Indian Penal Code, the major criminal law in India. 
The dataset and our codes are available at \url{https://github.com/Law-AI/LeSICiN}.

\section{Related Work} \label{sec:related}
Legal Statute Identification (LSI) has been widely studied by researchers in an attempt to automate the process. 

\noindent {\bf Initial Efforts:} The earliest LSI approaches used statistical algorithms along with hand-crafted rules~\cite{kort1957predicting,ulmer1963quantitative,segal1984predicting,lauderdale2012supreme}. 
Later, LSI was approached as a text classification problem based on manually engineered features~\cite{liu2006exploring,aletras2016predicting}.
However, such hand-crafted rules/features do not allow these methods to generalize well.

\noindent {\bf Neural models for LSI:} 
Recently developed attention-based neural models for LSI look to extract richer features from text using techniques such as dynamic context vectors~\cite{luo2017learning}, dynamic thresholding~\cite{wang2018modeling}, hierarchical classification~\cite{wang2019hierarchical} or pretrained BERT-based contextualizers~\cite{chalkidis2019neural}.
\citet{xu2020distinguish} split statutes into communities and use a novel graph distillation operator to identify intra-community features; however, they do not use the citation network in any way.
We use all the above-mentioned methods as baselines in this paper. 

\noindent {\bf LJP and Multi-task approaches:} Legal Judgment Prediction (LJP) is an umbrella problem that encompasses several related sub-tasks such as identifying statutes (LSI), charges, the term of penalty, etc. 
A different class of algorithms attempts to exploit the relationship between these multiple related  tasks (of LJP) by modeling them together~\cite{zhong2018legal,yang2019legal}. 
These models work effectively only when training data is available for more than one of the tasks.
Since our main focus is LSI, and we do not have training data for other tasks such as predicting the term of penalty, we do not consider these as baselines in this work.

\section{Data Preparation} \label{sec:data}

\label{sec:data}

We develop a novel dataset for the Legal Statute Identification (LSI) task using criminal case documents and statutes from the Indian judiciary. This Section describes the dataset.

\noindent {\bf Statutes:} In Indian Law, most criminal offences are described in the Indian Penal Code (IPC), which is an \textit{Act}.
The IPC {\it Act} has a hierarchical structure -- the \textit{Act} is divided into coarse-grained categories called \textit{Chapters}, which are further subdivided into fine-grained categories called \textit{Topics}. 
Each Topic groups together a set of {\it Sections} that are based on the same crime. Sections are statutory legal articles that are usually cited from case documents. 
The text of a Section describes the nature and circumstances of the crime, and litigation procedures involved.
In this paper, we use the terms `Statute' and `Section' interchangeably.
Table~\ref{tab:ipchier} shows a part of the Indian Penal Code, depicting examples of Chapters, Topics and Sections.

\begin{table}[tb]
\centering
\small
\begin{tabular}{|p{0.15\columnwidth}|p{0.23\columnwidth}|p{0.45\columnwidth}|}
\hline
\textbf{Chapter} & \textbf{Topic} & \textbf{Section} \\ \hline\hline

\multirow{4}{0.15\columnwidth}{Offences affecting Human Body} & \multirow{2}{0.23\columnwidth}{Offences affecting Life} & \textbf{299:} Culpable homicide \\ \cline{3-3}
& & \textbf{307:} Attempt to murder \\ \cline{2-3}
& \multirow{2}{0.2\columnwidth}{Hurt} & \textbf{321:} Voluntarily causing hurt \\ \cline{3-3}
& & \textbf{334:} Voluntarily causing hurt on provocation \\ \cline{1-3}
\multirow{4}{0.15\columnwidth}{Offences against Property} & \multirow{2}{0.2\columnwidth}{Robbery and Dacoity} & \textbf{390:} Robbery \\ \cline{3-3}
& & \textbf{396:} Dacoity with murder \\ \cline{2-3}
& \multirow{2}{0.2\columnwidth}{Criminal Trespass} & \textbf{441:} Criminal trespass \\ \cline{3-3}
& & \textbf{446:} House breaking by night \\ \hline
\end{tabular}%
\vspace{-2mm}
\caption{An illustrative part of the hierarchy of the Indian Penal Code (IPC) Act divided into Chapters, Topics and Sections (which are mostly cited by documents). 
}   
\label{tab:ipchier}
\vspace{-6mm}
\end{table}




\noindent {\bf Facts:} We collected $\sim 100K$ court case documents from the Supreme Court and six major High Courts in India, from the website \url{https://indiankanoon.org}.
We collected only documents that cite at least one Section from the IPC. 
Case documents are comprised of many semantic parts, such as the \textit{Facts}, \textit{arguments}, \textit{ruling}, etc.~\cite{bhattacharya2019identification}. 
Since the input to the LSI task consists of only the {\it Facts} (that led to filing of the case), 
we used the Hier-BiLSTM-CRF classifier developed by \citet{bhattacharya2019identification} to extract the Facts from the collected case documents (F1 of $0.839$ over Facts).

\noindent {\bf The ILSI dataset:}
The IPC contains more than $500$ Sections, but a large majority of them are seldom cited. Hence, we chose to focus on the $100$ most frequently cited Sections of IPC as the set of labels in our dataset. 
We consider the Facts from only those case documents that cite at least one of these top $100$ Sections, and we end up with $66,090$ such Facts.
Table~\ref{tab:dscmp} (last column) shows the basic statistics of the dataset developed in this work, which we call {\it Indian Legal Statute Identification} (ILSI) dataset.
Note that we mask named entities in the text; this is a common preprocessing step in Legal NLP, meant to minimize demographic bias in models~\cite{chalkidis2019neural}.

We split the dataset into  \textit{train}, \textit{validation} and \textit{test} parts in the ratio of $64:16:20$. 
Thus, we have $42,884$ training documents, $10,203$ validation documents, and $13,043$ test documents. 
Since this is a multi-label classification dataset, we ensure that the distribution of labels is balanced across all three sets, using {\it iterative stratification} \cite{sechidis2011stratification}.

Each input to the LSI task is the textual description of a Fact (obtained from a particular case document). The set of Sections cited in the said case document is considered the gold-standard set of labels for the given Fact. The ILSI dataset is available at  \url{https://github.com/Law-AI/LeSICiN}.

\begin{table}[tb]
\centering
\small
\begin{tabular}{|p{0.18\textwidth}|r|r|r|}
\hline
\textbf{Dataset} & \textbf{ECHR} & \textbf{CAIL} & \textbf{ILSI} \\ \hline

Language & English & Chinese & English \\ \hline
No. of documents/Facts & 11,478 & 2,676,075 & 66,090 \\ 
No. of labels/statutes & 66 & 183 & 100 \\ 
Avg. no. of words per doc & 2406 & 1444 & 1232 \\ 
{\bf Avg. no. of labels per doc in test set} & $0.71$ & $1.09$ & $\mathbf{3.78}$ \\ \hline
Statute/Label Text & No & Yes & Yes \\ \hline
\end{tabular}%
\vspace{-2mm}
\caption{Statistics of the new ILSI dataset and comparison with two other LSI datasets (ECHR and CAIL). ILSI reflects the multi-label nature of the LSI task more closely.}  
\label{tab:dscmp}
\vspace{-6mm}
\end{table}

\vspace{2mm}
\noindent {\bf Comparison of ILSI with existing LSI datasets:}
Table~\ref{tab:dscmp} compares ILSI with two popular datasets for the task of LSI, namely, ECHR~\cite{chalkidis2019neural} and CAIL~\cite{xiao2018cail2018}. 
The ECHR dataset is relatively small, and contains many documents which do not cite any statute/label (since the dataset was designed for both binary and multi-label classification). 
Hence, the average number of labels per document is less than $1$.
Also, notably, the ECHR dataset does {\it not contain any text for the statutes/labels.}
The CAIL dataset (in the Chinese language) is quite large and includes the text of the statutes.
But the average number of labels per document is very close to $1$ ($1.09$), indicating that a large fraction of documents do not cite more than one label.
Whereas, for ILSI, the average number of labels per document is relatively high ($3.78$), since a significant fraction of the documents cite more than one of the labels. 
Hence, the ILSI dataset reflects more truly the multi-label nature of the LSI problem.

\section{Formalization of the Problem} \label{sec:formal}

This Section discusses the standard formulation of the LSI task as a multi-label classification problem, and how we formulate LSI as a link prediction task over a graph.

\vspace{1mm}
\noindent {\bf Multi-label Classification Formulation:}
Let $F = \{f_1, f_2, \ldots, f_{|F|}\}$ denote the entire set of Fact descriptions (documents), 
and $S = \{s_1, s_2, \ldots, s_{|S|}\}$ denote the set of IPC Sections (labels). 
Since more than one Sections may be relevant to a Fact $f$, each instance is denoted as a tuple $\langle f, \mathbf y_f \rangle$.
Here, $\mathbf y_f \in \{0,1\}^{|S|}$, where $\mathbf y_f[s] \in \{0,1\}$ indicates whether Section $s$ is relevant to Fact $f$.

The LSI task requires us to develop a function $\mathcal{F(\cdot)}$ such that
$\mathcal{F}\left(f, S\right) = \hat{\mathbf y_f}$; 
where $\hat{\mathbf y_f} \in \{0,1\}^{|S|}$, with $\hat{\mathbf y_f}[s] \in \{0,1\}$ denoting the function's prediction of whether Section $s$ is relevant to Fact $f$.



\vspace{1mm}
\noindent {\bf Link Prediction Formulation:}
We model the LSI task using a graph formalism, over a {\sl legal citation network}.
Each Section and each Fact (from training instances) is treated as a node in a heterogeneous graph $\mathcal{G} = (\mathcal{V}, \mathcal{E})$, with a node mapping function $\phi: \mathcal{V} \rightarrow \mathcal{A}$ and an edge mapping function $\psi: \mathcal{E} \rightarrow \mathcal{R}$, where 
$\mathcal{A}$ and $\mathcal{R}$ represent the types of nodes and types of relations between nodes, respectively.
In our case, we have Fact nodes ($F$) and Section nodes ($S$), which are accompanied with their attributes.
Additionally, we have placeholder node types to denote the hierarchical levels of IPC -- Topics ($T$), Chapters ($C$) and  IPC Act ($A$) (see Section~\ref{sec:data}). 
There exist several types of relations between the nodes -- `cites' ($ct$) and `cited by' ($ctb$) relationships between nodes of types $F$ and $S$, `includes' ($inc$) and `part of' ($po$) relationships between the successive node hierarchy types $A$, $C$, $T$ and $S$. Figure~\ref{fig:arch}
shows a pictorial representation of this network.

We construct the network using the following steps.
(i)~Assign the set of nodes in each node type, e.g., $A$, $C$, $T$, $S$, $F$.  
Set $\mathcal{V} = \mathcal{V}_F \cup \mathcal{V}_S \cup \mathcal{V}_T \cup \mathcal{V}_C \cup \mathcal{V}_A$.
(ii) For given nodes $u$ and $v$ belonging to \textit{different, successive levels of hierarchy} in the Act (IPC)  s.t. $\phi(u) \in \{A, C, T\}$ and $\phi(v) \in \{C, T, S\}$, include links $(u, v) \in \mathcal{E}_\text{inc}$ and $(v, u) \in \mathcal{E}_\text{po}$ iff $v$ is categorized under the broader hierarchy level of $u$.
(iii)~For given nodes $u$ and $v$ s.t. $u \in \mathcal{V}_F$ and $v \in \mathcal{V}_S$, include links $(u, v) \in \mathcal{E}_\text{ct}$ (`cites' link) and $(v, u) \in \mathcal{E}_\text{ctb}$ (`cited by' link) iff $\mathbf y_u[v] = 1$.

Given a new Fact $f$ {\it at test time}, the task is to identify the relevant Sections. This problem can be seen as \textbf{inductive link prediction} over the network described above, i.e., predicting the possibility of the existence of a link between the new Fact $f$ and the Section nodes in the network. 
Note that inductive link prediction is the task of predicting a link between a pair of nodes, where {\it one or more nodes might not be visible to the model during training.}

Formally, modeling LSI as Link Prediction requires us to develop a function $\mathcal{Q}(.)$ such that $\mathcal{Q}(u,v, \mathcal{G}) = \hat{y_{uv}}$, where $u \in \mathcal{V}_F$, $v \in \mathcal{V}_S$ and $\hat{y_{uv}} \in \{0, 1\}$ denotes the function's prediction of whether a link should exist between $u$ and $v$.

\section{Proposed LSI Model: \model{}} \label{sec:method}

\begin{figure*}[t]
\centering
\includegraphics[width=0.8\textwidth,height=7cm]{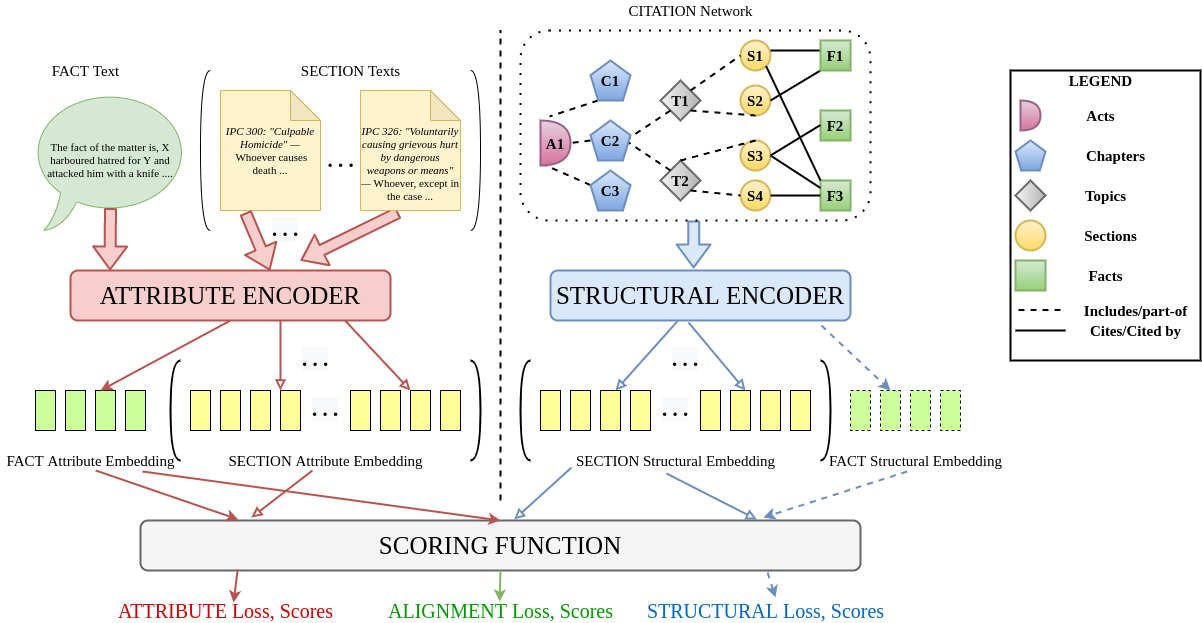} 
\caption{Architecture of our proposed model \textbf{\model}. The {\bf attribute encoder} is used for converting text to vector representations, and the {\bf structural encoder} is used to generate representations for Sections and training documents. These representations are combined and fed to the {\bf scoring mechanism} to generate losses and scores.}
\label{fig:arch}
\vspace{-6mm}
\end{figure*}

Figure~\ref{fig:arch} gives an overview of our proposed model \textbf{\model}. Each Fact $f \in F$ and Section $s \in S$ is accompanied by a text $x_f$ and $x_s$ respectively, which can be considered as attributes of the node. The citation network provides the structural information. We have two separate encoders for encoding attributes and structural information, which are shared across both Facts and Sections.

We adopt the technique used by DEAL~\cite{hao2020inductive} to generate good quality embeddings for unseen nodes (new Fact descriptions) using only their attributes at test time. 
During training, we obtain both attribute and structural embeddings of both node types, Facts and Sections. These are combined in different ways to generate three types of scores -- attribute, structural and alignment. During testing, since structural embeddings of new Facts are not available, we can only generate attribute and alignment scores. The alignment score enables the model to correlate the two kinds of embeddings, for generalizing well to unseen nodes during testing. 

Next, we describe the attribute and structural encoders, and the scoring function used by the architecture.



\subsection{Attribute Encoder} \label{sec:method-ae}
We use a shared attribute encoder for both Facts and Sections. A text portion in our formulation is a nested sequence of sentences and words. 
Hence we encode the text using the Hierarchical Attention Network (HAN)~\cite{yang2016hierarchical}.
Specifically, by feeding the Fact text $x_f$ and each Section text $x_s$ to HAN, we obtain a single, attention-weighted representation for the entire texts namely, $\mathbf h_f^{(a)}$ and a set of Section representations $\{\mathbf h_{s}^{(a)} \mid s \in S\}$.

\subsection{Structural Encoder} \label{sec:method-se}

To exploit the rich, semantic information of the Citation Network, we carefully design various metapath schemas, following the process adopted by \citet{fu2020magnn}. A {\sl metapath} is a sequence $A_1 \xrightarrow{R_1} A_2 \xrightarrow{R_2} \ldots \xrightarrow{R_l} A_{l+1}$, which denotes the composite relation $R_1 \circ R_2 \circ \ldots \circ R_l$ between node types $A_1, A_2, \ldots, A_{l+1}$. 
A metapath only defines a schema; there may be multiple sequence of nodes originating from the same starting node, following the same metapath schema $P$.
Each such sequence is called a metapath instance of $P$. The $P$ metapath-based neighbourhood of a node $v$, denoted as $\mathcal{N}_v^P$ is defined as all the nodes that can be reached from $v$ using the schema $P$. 
Any neighbour connected by two or more metapath instances is represented as two or more different nodes in $\mathcal{N}_v^P$.

To extract the structural information from the citation network, we make use of different metapath schemas with different node types as the start nodes. 
For example, for nodes of type $F$ (Facts) we have a metapath schema $F \xrightarrow{ct} S \xrightarrow{ctb} F$, capturing the type of relationship that exists between {\it Facts that cite the same Section}. From Figure~\ref{fig:arch}, we can observe multiple instances of this metapath schema, such as $F1-S1-F3$ and $F2-S3-F3$. 
As another example, for nodes of type $S$ (Sections), we have a metapath schema $S \xrightarrow{po} T \xrightarrow{po} C \xrightarrow{inc} T \xrightarrow{inc} S$, capturing the relationship between {\it Sections defined under the same Chapter}. 
From Figure~\ref{fig:arch}, we can see that $S1-T1-C2-T2-S3$ and $S1-T1-C2-T1-S2$ are two instances of this schema. 
We use a total of 4 Fact-side and 4 Section-side metapath schemas, which are listed in the {\bf supplementary material}.

\vspace{1mm}
\noindent Next, we describe the individual node representation, followed by intra- and inter-metapath aggregation to obtain the structural embedding.

\noindent\textbf{Node Embedding:} 
We have a set of parametric node embedding matrices $\{\mathbf{X}_A \mid A \in \mathcal{A}\}$ that are used to initialize each node $v$ with its feature vector $\mathbf x_v$.
Since the initial feature vectors might be of different dimensions and may map to different latent spaces, we need to transform them to the same dimensionality and space. 
For a node $v \in \mathcal{V}_A$ for $A \in \mathcal{A}$, we have 
$\mathbf x_v = \mathbf X_A \cdot \mathbf I_v$ and $\mathbf{h'}_v = \mathbf W_A \cdot \mathbf x_v$. 
Here $\mathbf X_A \in  \mathbb{R}^{d_A \times |\mathcal{V}_A|}$ is the node embedding matrix for nodes of type $A \in \mathcal{A}$;
$\mathbf I_v \in \mathbb{R}^{|\mathcal{V}_A|}$ is the one-hot identity vector for node $v$; and
$\mathbf x_v \in \mathbb{R}^{d_A}$ is the initial feature vector for $v$.
Further, $\mathbf W_A \in \mathbb{R}^{d' \times d_A}$ is the parametric transformation matrix for node type $A \in \mathcal{A}$; and
$\mathbf{h'}_v \in \mathbb{R}^{d'}$ is the transformed latent vector of uniform dimensionality (across node types).
After transformation, all node embeddings are ready to be fed to the aggregation architecture.

\noindent\textbf{Intra-Metapath Aggregation:} First, we need to aggregate all the node embeddings for a target node $v$ under a particular metapath schema $P$. For a metapath instance $P(v,u)$ connecting $v$ with its metapath-based neighbour $u \in \mathcal{N}_v^P$, we use a metapath instance encoder $g_\theta(.)$ to generate a representation for the instance $P(v,u)$. We adopt the relational rotation encoder used by \citet{fu2020magnn}.

Consider that $P(v,u) = \{n_0, n_1, \ldots, n_M\}$, with $n_0 = u$ and $n_M = v$. We thus have
\[\mathbf q_i = \mathbf{h'}_{n_i} + \mathbf q_{i - 1} \odot \mathbf r_i; \quad \mathbf h_{P(v,u)} = \frac{\mathbf q_M}{M + 1} \]
where $\mathbf q_0 = \mathbf{h'}_u$, $\mathbf r_i \in \mathbb{R}^{d'}$ is the learned relation vector for $R_i \in \mathcal{R}$ and $\mathbf h_{P(u,v)}$ is the vector representation of the metapath instance $P(u,v)$.  
Now, we have obtained representations $\mathbf h_{P(u,v)}$ for each $u \in \mathcal{N}_v^P$. We attentively combine these $P$-based representations for the target node $v$ as 
\[e_{vu}^P = \textsc{LeakyReLU} \left( \mathbf a_P^\intercal \cdot [\mathbf{h'}_v || \mathbf h_{P(v,u)}] \right)\]
%
\[\alpha_{vu}^P = \textsc{Softmax}_{u \in \mathcal{N}_v^P} \left( e_{vu}^P \right)\]
\[\mathbf h_v^P = \textsc{ReLU} \left( \sum_{u \in \mathcal{N}_v^P} \alpha_{vu}^P \cdot \mathbf h_{P(v,u)} \right)\]
where $\mathbf a_P \in \mathbb{R}^{2d'}$ is the parameterized context vector and $\mathbf h_v^P \in \mathbb{R}^{d'}$ is the aggregated representation from all metapath neighbours of $v$ in $P$.

\noindent\textbf{Inter-Metapath Aggregation:} Now, we need to aggregate the metapaths across different schemas. Consider $\mathcal{P}_A = \{P_1, P_2, \ldots, P_N\}$ as the set of metapath schemas which start with node type $A \in \mathcal{A}$. For a node $v \in \mathcal{V}_A$ of type $A$, we have a set of $N$ representations $\{\mathbf h_v^{P_i}, \forall i \in [1,N]\}$.

First, information for each metapath schema is summarized across all nodes $v \in \mathcal{V}_A$ as 
\[\mathbf s_{P_i} = \frac{1}{|\mathcal{V}_A|} \sum\limits_{v \in \mathcal{V}_A} \text{tanh} \left( \mathbf M_A \cdot \mathbf h_v^{P_i} + \mathbf b_A \right)\]
where $\mathbf M_A \in \mathbb{R}^{d_m \times d'}$ and $\mathbf b_A \in \mathbb{R}^{d_m}$ are learnable parameters.
Then, attention mechanism is employed to aggregate all the $M$ representations as $e_{P_i} = \mathbf q_A^\top \cdot \mathbf s_{P_i}$,
%
\[\beta_{P_i} = \textsc{Softmax}_{P_i \in \mathcal{P}_A} \left(e_{P_i}\right) \quad \mathbf h_v = \sum\limits_{P_i \in \mathcal{P}_A} \beta_{P_i} \mathbf h_v^{P_i}\]
where $\mathbf q_A \in \mathbb{R}^{d_m}$ is the parameterized context vector.   

Thus, at the end of the structural encoding phase, we get a single representation of the Fact $\mathbf h_f^{(s)}$ and a set of representations for each Section $\{\mathbf h_{s}^{(s)} \mid s \in S\}$.

\subsection{Scoring Mechanism} \label{sec:method-sm}
We design a scoring function $m_\theta(f; \{s \mid s \in S\})$ to assign a score to each Section $s \in S$ for Fact $f$. 
Leveraging the Fact that Sections in IPC do have a defined sequential order and related semantics, we first use an LSTM to contextualize these embeddings $\mathbf h_s$ for $s \in S$ as 
${\mathbf{\tilde{h}}_s} = \textsc{Bi-LSTM} \left( \mathbf h_s ; [\mathbf h_t \mid t \in S]\right)$.
Then, we use the standard attention mechanism to generate a single aggregated embedding $\mathbf h_S$ representing the set $S$:
\[ e_{s} = \mathbf w_S^\top \cdot \text{tanh} \left( \mathbf M_S \cdot \mathbf{\tilde{h}_s} + \mathbf b_S \right) \]
\[\gamma_{s} = \textsc{Softmax}_{s \in S} \left(e_s\right) \quad \mathbf h_S = \sum\limits_{s \in S} \gamma_{s} \mathbf{\tilde{h}}_s\]
where $\mathbf w_S \in \mathbb{R}^{d_s}$ is the parameterized context vector and $\mathbf M_S \in \mathbb{R}^{d_s \times d'}$ and $\mathbf b_S \in \mathbb{R}^{d_s}$ are learnable parameters.

Finally, we generate the score for each class as:
\[\mathbf o_f = m_\theta(f; \{s \mid s \in S\}) = \sigma \left( \mathbf W_C \cdot [\mathbf h_f || \mathbf h_S] + \mathbf b_C \right) \]
where $\mathbf W_C \in \mathbb{R}^{|S| \times 2d'}$ and $\mathbf b_C \in \mathbb{R}^{|S|}$ are the learnable parameters for the final classification layer.

Since we have two sets of embeddings for each Fact $f$ and Section $c_i$, we can generate three sets of scores, namely -- 

\noindent (i) \textbf{Attribute Score:} $\mathbf o_f^{(a)} = m_\theta(\mathbf h_f^{(a)}; \{\mathbf h_s^{(a)} \mid s \in S\})$ matches the attribute embedding of Facts with Sections; 

\noindent (i) \textbf{Structural Score:} $\mathbf o_f^{(s)} = m_\theta(\mathbf h_f^{(s)}; \{\mathbf h_s^{(s)} \mid s \in S\})$ matches the structural embedding of Facts with Sections; 

\noindent (iii) \textbf{Alignment Score:} $\mathbf o_f^{(l)} = m_\theta(\mathbf h_f^{(a)}; \{\mathbf h_s^{(s)} \mid s \in S\})$ matches the attribute embedding of Facts with structural embedding of Sections.

The structural score can only be calculated during training time, since the graphical structure of the Facts at test/inference time is not available. During training, the structural score helps the model to understand the graphical structure of Sections and training documents.
However, the graphical structure of the Sections is available at all times, and thus the alignment score actually tunes the model to generate similar attribute and structural representations.

\noindent \textbf{Using Dynamic Context:} To provide better guidance to the attention mechanism for the structural encoder through the attribute embeddings, we replace the static context vectors in the structural encoder with dynamically generated vectors from the attribute embeddings of the same node~\cite{luo2017learning}. In the scoring function, we use the Fact embeddings to generate the dynamic context.
\[ \mathbf a_P = \mathbf T_P \cdot \mathbf h_v^{(a)}; \quad \mathbf q_A = \mathbf T_A \cdot \mathbf h_v^{(a)}; \quad \mathbf w_S = \mathbf T_S \cdot \mathbf h_f\]
where $\mathbf T_P \in \mathbb{R}^{2d' \times d'}$, $\mathbf T_A \in \mathbb{R}^{d' \times d'}$ and $\mathbf T_S \in \mathbb{R}^{d' \times d'}$ are the learnable transformation matrices.


\subsection{Training and Prediction} \label{sec:method-tp}

The loss function has three parts, analogous to the three scores, namely, $\mathcal{L}^{(a)}$, $\mathcal{L}^{(s)}$ and $\mathcal{L}^{(l)}$. We use weighted Binary Cross Entropy Loss to calculate each component as:
\[l_s^{(t)} = w_s \mathbf y_f[s] \log{(\mathbf o_f^{(t)}[s])} + (1 - \mathbf y_f[s]) \log{(1 - \mathbf o_f^{(t)}[s])}\]
\[\mathcal{L}^{(t)} = -\frac{1}{|B|} \sum\limits_{f \in B} \sum\limits_{s \in S} l_s^{(t)}; \quad t \in \{a,s,l\}, s \in S\]
where $w_s$ denotes the weight of each positive sample of class $s \in S$ and $B$ denotes a mini-batch of Facts.

The {\it vanilla weighting scheme} (VWS) assigns $w_s = N/f_s, \forall s \in S$, where $N$ is the total no. of training documents and $f_s$ is the  no. of training documents that cite $s$.
However, this scheme can sometimes lead to very large weights for the rare labels $f_s << N$. 
To address this issue, we propose a {\it threshold-based weighting scheme} (TWS) defined as $w_s = \min{(f_{max} / f_s, \eta)}$
where $f_{max}$ is the frequency of the most cited label, and $\eta$ is a threshold value determined on the validation set.
This scheme caps the class weights at a reasonable value, so that the model does not compromise performance on the frequent classes for minor improvements over the rare ones. 


We have the final loss as:
$\mathcal{L} = \theta_a \mathcal{L}^{(a)} + \theta_s \mathcal{L}^{(s)} + \theta_l \mathcal{L}^{(l)}$
During test time, the structural score is not available. Thus,
$\hat{\mathbf y_f} = I \left(\lambda_a \mathbf o_f^{(a)} + \lambda_l \mathbf o_f^{(l)} \geq \tau \right)$ 
where $\hat{\mathbf y_f} \in \{0,1\}^{|S|}$ and $\hat{\mathbf y_f}[s] \in \{0,1\}$ indicates whether the model predicts Section $s$ to be relevant to Fact $f$ or not, and $\tau$ is the threshold value set using the validation set. 

\section{Experiments \& Results} \label{sec:expt}

In this Section, we compare the performance of our model with that of several baselines. We also analyze the effectiveness of our model and its various components.


\begin{table*}[!t]
\centering
\small
\begin{tabular}{|p{0.1\textwidth}|rrrr|rr|r|rr|}
\hline
\textbf{Metric} & \textbf{FLA} & \textbf{DPAM} & \textbf{HMN} & \textbf{LADAN} & \textbf{HBERT} & \textbf{HLegalBERT} & \textbf{DEAL} & \textbf{\model} & \textbf{\model{} ($\tau = 0.3$)} \\ \hline
Macro-P & $12.19$ & $27.11$ & $10.27$ & $12.54$ & $5.79$ & $4.36$ & $12.66$ & $\mathbf{27.90}$ & $11.99$\\ 
Macro-R & $69.25$ & $27.43$ & $57.30$ & $46.17$ & $51.43$ & $52.55$ & $64.83$ & $31.32$ & $\mathbf{69.42}$ \\
Macro-F1 & $19.60$ & $23.86$ & $16.12$ & $17.93$ & $7.91$ & $7.58$ & $19.43$ & $\mathbf{28.45}$ & $19.77$ \\
Jaccard & $11.64$ & $14.44$ & $9.56$ & $10.28$ & $4.47$ & $4.01$ & $11.44$ & $\mathbf{17.57}$ & $11.49$ \\ 
\hline 
\end{tabular}%
\vspace{-3mm}
\caption{Comparative results of the baselines and the proposed \model{} on the test set. Last column shows a variations of \model{}, that optimizes for Recall. All values are in percentages. Best value for each metric is in boldface. Differences in Macro-F1 between \model{} and all baselines are all statistically significant (paired t-Test with 95\% confidence).}   
\label{tab:res}
\vspace{-5mm}
\end{table*}

\vspace{2mm}
\noindent {\bf Baselines:}
We consider the following state-of-the-art baselines for ILSI:
\textbf{(1)~FLA}~\cite{luo2017learning} uses two HANs to generate Fact embeddings 
and Section embeddings 
\textbf{(2)~DPAM}~\cite{wang2018modeling} learns the distribution of {\it pairs of Sections} instead of individual Sections, for better learning the rare Sections; 
\textbf{(3)~HMN}~\cite{wang2019hierarchical} models the task as a hierarchical classification task with two levels (analogous to \textit{Acts} and \textit{Sections} in our work); 
\textbf{(4)~LADAN}~\cite{xu2020distinguish} groups the Sections into non-overlapping communities based on TF-IDF similarity measures before performing Graph distillation on each community
\textbf{(5)~HBERT}~\cite{chalkidis2019neural} uses BERT~\cite{devlin2018bert} to generate embeddings for each sentence of the Fact, and an LSTM-Attn layer on top to generate the final embedding for each document before classifying. Note that this model {\it does not utilize the text of the statutes};
\textbf{(6)~HLegalBERT:} A variation of HBERT with the same architecture, but using LegalBERT~\cite{legalbert} to generate domain-aware embeddings for each sentence;
\textbf{(7)~DEAL:} The base DEAL~\cite{hao2020inductive} using a node embedding matrix as the structural encoder, cosine similarity for scoring links, and a network distance-based weighting of the negative samples for the loss function.



\vspace{2mm}
\noindent {\bf Hyperparameters:}
We apply the same settings to all competing methods to ensure fair competition. 
Every model is trained and validated on the train and validation sets described in Section~\ref{sec:data}, for $100$ epochs. 
At the end of training, the model state yielding the best validation result is used over the test set. 
Some hyper-parameters are also fixed across all models -- 
(i)~embedding dimension of $200$ in all cases except $768$ for HBERT, 
(ii)~Adam Optimizer with learning rate in the range $[0.01, 0.000001]$, 
(iii)~dropout probability of $0.5$, and 
(iv)~batch size of $32$.

For model-specific configurations required in \model, we sample 8 metapath instances per schema, per node. To place extra emphasis on network learning, we set $\theta_a = 1$, $\theta_s = 2$, $\theta_l = 3$ and $\lambda_a = 0.25$, $\lambda_l = 0.75$. Based on validation set performance, we set $\tau = 0.65$ and $\eta = 10.0$. 

To improve the models' understanding of legal text, we use sent2vec~\cite{sent2vec} embeddings of dimension $200$ trained on a large legal corpus of Indian case documents, to initialize the sentence embeddings of all models (except HBERT and HLegalBERT which have their own encoding methods).

We have used the source codes for the baselines where available. We have coded our model and other models (if code unavailable) in PyTorch~\cite{paszke2019pytorch}. For the network part, we use PyTorch Geometric \cite{fey2019fast}.
We trained all models on an NVIDIA Tesla T4 with 16 GB GPU memory. 

\vspace{2mm}
\noindent {\bf Evaluation Metrics:}
Since the label distribution (frequency of statutes being cited) is very skewed, 
we use macro-averaged versions  (i)~\textit{macro-Precision}, (ii)~\textit{macro-Recall} and (iii)~\textit{macro-F1}. Also, we use (iv)~the Jaccard overlap between the gold standard set of labels and the predicted set of labels (of a test Fact). All metrics are averaged over the entire test dataset.


\subsection{Comparative Results} \label{sub:comparative}

Table~\ref{tab:res} shows the performance of all the competing models on the test set. In terms of macro-F1, our proposed model \model{} performs the best across all methods, followed by DPAM and FLA.
The last column of Table~\ref{tab:res} shows a configuration of \model{} tuned for higher recall, by reducing the classification threshold to $\tau = 0.3$. This is for the sake of comparison with the best configuration of \model{} (second-last column), which has been tuned for the highest F1 with $\tau = 0.65$.


In the case of FLA, the Fact-aware Section representations seem to give good performance, especially in terms of Recall (highest macro-Recall across all methods).
Methods like HMN and LADAN that perform two-step classification do not perform well,
possibly since performance of the second step depends crucially on that of the first step; the distribution of Sections over Topics (in case of HMN) and Sections over communities (for LADAN) might be too skewed, leading to low Precision values (and hence low F-score) for both models.
The BERT-based HBERT and HLegalBERT do not perform well at all, most possibly because these models do not make use of the Section texts.  \footnote{We follow the HBERT implementation in~\cite{chalkidis2019neural}.} 
DEAL uses the graph structure to extract valuable distinguishing features, achieving the second highest macro-Recall after FLA. 
However, it does not make use of the graph heterogeneity nor complex GNNs, just a simple node embedding lookup. Thus its overall performance (macro-F1) is relatively low.

Interestingly, several models suffer from the problem of low Precision and high Recall, demonstrating a significant difference between macro-P and macro-R, except for DPAM and \model, which are actually the top performing methods. 
However, as the last column in Table~\ref{tab:res} shows, \model{} can be tuned to yield higher recall than all baselines with slight sacrifice in F1. \model{} with $\tau = 0.3$ performs similar to FLA in terms of all the metrics.  

\subsection{Analysis of Our Model}

\noindent {\bf Ablation experiments:}
To understand the effectiveness of different components of \model, we try the following variations:

\noindent (1) \textit{\model-E:} To inspect the graph encoding capabilities of \model{}, we replace the metapath-based aggregation network with a simple node embedding lookup table, similar to what is used in DEAL. 

\noindent (2) \textit{\model-B:} We replace the Scoring Function described in Section~\ref{sec:method-sm} with a simple Bilinear layer, i.e., for each $\langle Fact, Sec \rangle$ pair, this variation generates a single score that denotes how relevant the members of the pair are.

\noindent (3) \textit{\model-S:} We remove the structural loss $\mathcal{L}^{(s)}$ calculated based on Section structural embeddings and Fact attribute embeddings during training.

\noindent (4) \textit{\model-V:} We use VWS for weighting loss instead of our proposed TWS (described in Section~\ref{sec:method-tp}).

\noindent (5) \textit{\model-SG:} To investigate the gains of modeling heterogeneity in the citation network, we replace the metapath-based aggregation network with a homogeneous graph encoder, GraphSAGE \cite{hamilton2017inductive}.

\begin{table}[tb]
\centering
\small
\begin{tabular}{|p{0.1\textwidth}|r|r|r|r|}
\hline
\textbf{Model} & \textbf{Macro-P} & \textbf{Macro-R} & \textbf{Macro-F1} & \textbf{Jacc} \\ \hline

\model & $27.90$ & $31.32$ & $28.45$ & $17.57$ \\ \hline
\model-E & $27.87$ & $26.99$ & $25.18$ & $15.38$ \\ 
\model-B & $21.53$ & $36.04$ & $24.66$ & $14.89$ \\
\model-S & $22.00$ & $43.19$ & $25.40$ & $15.32$ \\
\model-V & $19.50$ & $47.60$ & $25.52$ & $15.48$  \\ 
\model-SG & $26.50$ & $29.80$ & $26.90$ & $16.17$ \\
\hline
\end{tabular}%
\vspace{-3mm}
\caption{Ablation test results on our proposed model \model{} on the test set. All values are in percentages.}   
\label{tab:abl}
\vspace{-6mm}
\end{table}

\vspace{1mm}
\noindent Table~\ref{tab:abl} shows that removal of any component from \model{} degrades its performance. 
The node embedding matrix of \model-E is not capable of capturing the complex semantic relationships in the heterogeneous citation graph and hence fails to generalize well to the test dataset.
The scoring function used in \model~seems to have a higher impact on performance, since replacing it with a bilinear matching function (in \model-B) deteriorates macro-F1 even further. 
The bilinear function cannot model relationships between labels efficiently since it scores each label independently.
\model-S, though inferior to \model, outperforms several baselines, probably due to the fact that the alignment loss captures the Section-side aspect of the citation network.
Using threshold-based weights also seem to improve performance, since \model{} outperforms \model-V with the vanilla weighting scheme.
Finally, \model-SG does not perform as good as \model{} since it neglects the heterogeneity of the citation network.
Importantly, despite significant ablations, the variations of \model{} achieve higher macro-F1 than the baselines in Table~\ref{tab:res}.

\begin{figure}[tb]
\centering
\begin{subfigure}[b]{0.49\columnwidth}
\centering
\includegraphics[width=\textwidth,height=3cm]{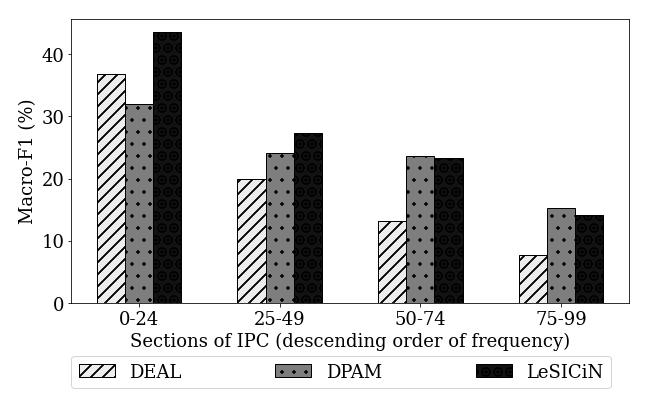}
\caption{Label frequency}
\label{fig:labanal}
\end{subfigure}
\begin{subfigure}[b]{0.49\columnwidth}
\centering
\includegraphics[width=\textwidth,height=3cm]{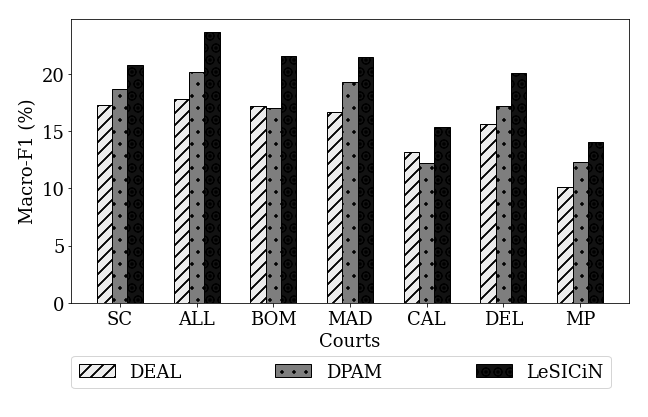}
\caption{Across Law courts}
\label{fig:crt}
\end{subfigure}
\vspace{-3mm}
\caption{Performance (macro-F1) of \model{} and two closest competitors DPAM and DEAL across (a)~labels (statutes) of varying frequencies, (b)~Facts from different Law courts. 
}
\label{fig:user_representativeness}
\vspace{-6mm}
\end{figure}

\vspace{1mm}
\noindent {\bf Performance on labels of varying frequencies:}
We check how well \model{} and the two closest baselines (DPAM and DEAL) perform 
across labels/Sections having different frequencies of citation.
Figure~\ref{fig:labanal} shows the performances (Macro-F1 scores), where the X-axis shows the Sections sorted in decreasing order of their frequency of being cited (by the documents in the test set), and grouped into four groups.
The performances of all models degrade sharply over the less frequently cited labels/Sections.
\model{} performs much better than the baselines for the first two groups of Sections, i.e., for the $50$ most-frequently cited Sections.
For the less frequently cited Sections, \model{} performs much better than DEAL and very close to DPAM (difference between \model{} and DPAM is {\it not} statistically significant by paired t-Test with 95\% confidence).




\vspace{1mm}
\noindent {\bf Performance on Facts from different courts of Law:}
As discussed in Section~\ref{sec:data}, the ILSI dataset has Facts extracted from case documents of {\it seven} Indian courts: the Supreme Court and six major High Courts (Allahabad, Bombay, Madras, Calcutta, Delhi and Madhya Pradesh). 
It is important for models to be  generalizable across different courts, since writing styles may vary significantly across courts.





Figure~\ref{fig:crt} shows the performance of \model{} and the two closest competitors (DPAM and DEAL) over the Facts of each Court. \model{} consistently performs much better than both DPAM and DEAL, across documents from all courts, thus showing its generalizability over writing styles. 

\section{Conclusion}
We propose a novel model \model{} for the Legal Statute Identification (LSI) task, that utilizes the citation network of the training data to produce robust, feature-rich representations for the text, which generalize well to unseen test documents, across a range of different parameters.
We also introduce a new, large-scale English dataset for LSI constructed from Indian judiciary documents. 


\vspace{2mm}
\noindent {\bf Acknowledgements:}
The research is partially supported by TCG CREST through the project ``Smart Legal Consultant: AI-based Legal Analytics''.
S. Paul is supported by the Prime Minister's Research Fellowship (PMRF) from the Government of India.

\bibliography{aaai22}

\begin{thebibliography}{26}
\providecommand{\natexlab}[1]{#1}

\bibitem[{Aletras et~al.(2016)Aletras, Tsarapatsanis, Preo{\c{t}}iuc-Pietro,
  and Lampos}]{aletras2016predicting}
Aletras, N.; Tsarapatsanis, D.; Preo{\c{t}}iuc-Pietro, D.; and Lampos, V. 2016.
\newblock Predicting judicial decisions of the European Court of Human Rights:
  A natural language processing perspective.
\newblock \emph{PeerJ Computer Science}, 2: e93.

\bibitem[{Bhattacharya et~al.(2020)Bhattacharya, Ghosh, Pal, and
  Ghosh}]{hier-spc-net}
Bhattacharya, P.; Ghosh, K.; Pal, A.; and Ghosh, S. 2020.
\newblock Hier-SPCNet: A Legal Statute Hierarchy-Based Heterogeneous Network
  for Computing Legal Case Document Similarity.
\newblock In \emph{Proc. ACM SIGIR Conference on Research and Development in
  Information Retrieval}, 1657–1660.

\bibitem[{Bhattacharya et~al.(2019)Bhattacharya, Paul, Ghosh, Ghosh, and
  Wyner}]{bhattacharya2019identification}
Bhattacharya, P.; Paul, S.; Ghosh, K.; Ghosh, S.; and Wyner, A. 2019.
\newblock Identification of Rhetorical Roles of Sentences in Indian Legal
  Judgments.
\newblock In \emph{Proceedings of JURIX}.

\bibitem[{Chalkidis, Androutsopoulos, and Aletras(2019)}]{chalkidis2019neural}
Chalkidis, I.; Androutsopoulos, I.; and Aletras, N. 2019.
\newblock Neural Legal Judgment Prediction in English.
\newblock In \emph{Proceedings of ACL}.

\bibitem[{Chalkidis et~al.(2020)Chalkidis, Fergadiotis, Malakasiotis, Aletras,
  and Androutsopoulos}]{legalbert}
Chalkidis, I.; Fergadiotis, M.; Malakasiotis, P.; Aletras, N.; and
  Androutsopoulos, I. 2020.
\newblock {LEGAL}-{BERT}: The Muppets straight out of Law School.
\newblock In \emph{Findings of the Association for Computational Linguistics:
  EMNLP 2020}, 2898--2904.

\bibitem[{Devlin et~al.(2018)Devlin, Chang, Lee, and
  Toutanova}]{devlin2018bert}
Devlin, J.; Chang, M.-W.; Lee, K.; and Toutanova, K. 2018.
\newblock Bert: Pre-training of deep bidirectional transformers for language
  understanding.
\newblock In \emph{Proceedings of NAACL}.

\bibitem[{Fey and Lenssen(2019)}]{fey2019fast}
Fey, M.; and Lenssen, J.~E. 2019.
\newblock Fast Graph Representation Learning with {PyTorch Geometric}.
\newblock In \emph{ICLR Workshop on Representation Learning on Graphs and
  Manifolds}.

\bibitem[{Fu et~al.(2020)Fu, Zhang, Meng, and King}]{fu2020magnn}
Fu, X.; Zhang, J.; Meng, Z.; and King, I. 2020.
\newblock Magnn: Metapath aggregated graph neural network for heterogeneous
  graph embedding.
\newblock In \emph{Proceedings of The Web Conference 2020}, 2331--2341.

\bibitem[{Hamilton et~al.(2017)}]{hamilton2017inductive}
Hamilton, W.~L.; et~al. 2017.
\newblock Inductive Representation Learning on Large Graphs.
\newblock NIPS'17.

\bibitem[{Hao et~al.(2020)Hao, Cao, Fang, Xie, and Wang}]{hao2020inductive}
Hao, Y.; Cao, X.; Fang, Y.; Xie, X.; and Wang, S. 2020.
\newblock Inductive Link Prediction for Nodes Having Only Attribute
  Information.
\newblock In Bessiere, C., ed., \emph{Proceedings of the Twenty-Ninth
  International Joint Conference on Artificial Intelligence, {IJCAI-20}},
  1209--1215. International Joint Conferences on Artificial Intelligence
  Organization.

\bibitem[{Kort(1957)}]{kort1957predicting}
Kort, F. 1957.
\newblock Predicting Supreme Court decisions mathematically: A quantitative
  analysis of the “right to counsel” cases.
\newblock \emph{American Political Science Review}, 51(1): 1--12.

\bibitem[{Lauderdale and Clark(2012)}]{lauderdale2012supreme}
Lauderdale, B.~E.; and Clark, T.~S. 2012.
\newblock The Supreme Court's many median justices.
\newblock \emph{American Political Science Review}, 106(4): 847--866.

\bibitem[{Liu and Hsieh(2006)}]{liu2006exploring}
Liu, C.-L.; and Hsieh, C.-D. 2006.
\newblock Exploring phrase-based classification of judicial documents for
  criminal charges in chinese.
\newblock In \emph{International Symposium on Methodologies for Intelligent
  Systems}, 681--690. Springer.

\bibitem[{Luo et~al.(2017)Luo, Feng, Xu, Zhang, and Zhao}]{luo2017learning}
Luo, B.; Feng, Y.; Xu, J.; Zhang, X.; and Zhao, D. 2017.
\newblock Learning to predict charges for criminal cases with legal basis.
\newblock In \emph{Proceedings of EMNLP}, 2727--2736.

\bibitem[{Pagliardini, Gupta, and Jaggi(2018)}]{sent2vec}
Pagliardini, M.; Gupta, P.; and Jaggi, M. 2018.
\newblock {Unsupervised Learning of Sentence Embeddings using Compositional
  n-Gram Features}.
\newblock In \emph{Proc. Conference of the North {A}merican Chapter of the
  Association for Computational Linguistics: Human Language Technologies
  (NAACL-HLT)}, 528--540.

\bibitem[{Paszke et~al.(2019)Paszke, Gross, Massa, Lerer, Bradbury, Chanan,
  Killeen, Lin, Gimelshein, Antiga, Desmaison, Kopf, Yang, DeVito, Raison,
  Tejani, Chilamkurthy, Steiner, Fang, Bai, and Chintala}]{paszke2019pytorch}
Paszke, A.; Gross, S.; Massa, F.; Lerer, A.; Bradbury, J.; Chanan, G.; Killeen,
  T.; Lin, Z.; Gimelshein, N.; Antiga, L.; Desmaison, A.; Kopf, A.; Yang, E.;
  DeVito, Z.; Raison, M.; Tejani, A.; Chilamkurthy, S.; Steiner, B.; Fang, L.;
  Bai, J.; and Chintala, S. 2019.
\newblock PyTorch: An Imperative Style, High-Performance Deep Learning Library.
\newblock In Wallach, H.; Larochelle, H.; Beygelzimer, A.; d\textquotesingle
  Alch\'{e}-Buc, F.; Fox, E.; and Garnett, R., eds., \emph{Advances in Neural
  Information Processing Systems 32}, 8024--8035. Curran Associates, Inc.

\bibitem[{Sechidis, Tsoumakas, and Vlahavas(2011)}]{sechidis2011stratification}
Sechidis, K.; Tsoumakas, G.; and Vlahavas, I. 2011.
\newblock On the stratification of multi-label data.
\newblock In \emph{Joint European Conference on Machine Learning and Knowledge
  Discovery in Databases}, 145--158. Springer.

\bibitem[{Segal(1984)}]{segal1984predicting}
Segal, J.~A. 1984.
\newblock Predicting Supreme Court cases probabilistically: The search and
  seizure cases, 1962-1981.
\newblock \emph{American Political Science Review}, 78(4): 891--900.

\bibitem[{Ulmer(1963)}]{ulmer1963quantitative}
Ulmer, S.~S. 1963.
\newblock Quantitative analysis of judicial processes: Some practical and
  theoretical applications.
\newblock \emph{Law \& Contemp. Probs.}, 28: 164.

\bibitem[{Wang et~al.(2019)Wang, Fan, Niu, Yang, Zhang, and
  Guo}]{wang2019hierarchical}
Wang, P.; Fan, Y.; Niu, S.; Yang, Z.; Zhang, Y.; and Guo, J. 2019.
\newblock Hierarchical Matching Network for Crime Classification.
\newblock In \emph{Proceedings of SIGIR}, 325--334. ACM.

\bibitem[{Wang et~al.(2018)Wang, Yang, Niu, Zhang, Zhang, and
  Niu}]{wang2018modeling}
Wang, P.; Yang, Z.; Niu, S.; Zhang, Y.; Zhang, L.; and Niu, S. 2018.
\newblock Modeling dynamic pairwise attention for crime classification over
  legal articles.
\newblock In \emph{Proceedings of SIGIR}, 485--494. ACM.

\bibitem[{Xiao et~al.(2018)Xiao, Zhong, Guo, Tu, Liu, Sun, Feng, Han, Hu, Wang
  et~al.}]{xiao2018cail2018}
Xiao, C.; Zhong, H.; Guo, Z.; Tu, C.; Liu, Z.; Sun, M.; Feng, Y.; Han, X.; Hu,
  Z.; Wang, H.; et~al. 2018.
\newblock Cail2018: A large-scale legal dataset for judgment prediction.
\newblock \emph{arXiv preprint arXiv:1807.02478}.

\bibitem[{Xu et~al.(2020)Xu, Wang, Chen, Pan, Wang, and
  Zhao}]{xu2020distinguish}
Xu, N.; Wang, P.; Chen, L.; Pan, L.; Wang, X.; and Zhao, J. 2020.
\newblock Distinguish Confusing Law Articles for Legal Judgment Prediction.
\newblock In \emph{Proceedings of ACL}.

\bibitem[{Yang et~al.(2019)Yang, Jia, Zhou, and Luo}]{yang2019legal}
Yang, W.; Jia, W.; Zhou, X.; and Luo, Y. 2019.
\newblock Legal judgment prediction via multi-perspective bi-feedback network.
\newblock In \emph{Proceedings of IJCAI}.

\bibitem[{Yang et~al.(2016)Yang, Yang, Dyer, He, Smola, and
  Hovy}]{yang2016hierarchical}
Yang, Z.; Yang, D.; Dyer, C.; He, X.; Smola, A.; and Hovy, E. 2016.
\newblock Hierarchical attention networks for document classification.
\newblock In \emph{Proceedings of NAACL}, 1480--1489.

\bibitem[{Zhong et~al.(2018)Zhong, Guo, Tu, Xiao, Liu, and
  Sun}]{zhong2018legal}
Zhong, H.; Guo, Z.; Tu, C.; Xiao, C.; Liu, Z.; and Sun, M. 2018.
\newblock Legal judgment prediction via topological learning.
\newblock In \emph{Proceedings of EMNLP}, 3540--3549.

\end{thebibliography}

\begin{appendices}

\section{Overview of Algorithm} \label{sec:algo}

\vspace{-3mm}
\begin{algorithm}[H] 
\caption{Forward Propagation of LeSICiN}
\label{alg:main}
\begin{algorithmic}[1]

\Require $x_f$ --- the text of the Facts

$\{x_s \mid s \in S\}$ --- the text of the Sections

$\mathcal G, \phi, \psi$ --- the Citation Network with node and edge mapping functions

training --- indicates if this is a training pass

\Ensure $\mathbf o_f$ --- Final predictions for each Section

$\mathcal L$ --- Final loss

\vspace{2mm}

\For {$s \in S$}
    \State Encoding Section Text: 
    
    $\mathbf h_s^{(a)} = \textsc{Attribute-Encoder}(x_s)$
\EndFor
\State Encoding Fact Text:

$\mathbf h_f^{(a)} = \textsc{Attribute-Encoder}(x_f)$

\State Generate Dynamic Context Vectors:

$\mathbf a_P = \mathbf T_P \cdot \mathbf h_v^{(a)}; \quad \mathbf q_A = \mathbf T_A \cdot \mathbf h_v^{(a)};$

$\quad \mathbf w_S = \mathbf T_S \cdot \mathbf h_f$

\For {$s \in S$}
    \State Encoding Section Structure: 
    
    $\mathbf h_s^{(s)} = \textsc{Structural-Encoder}(\mathcal G, \phi, \psi; \mathbf a_P, \mathbf q_A)$
\EndFor

\If {training {\bf is True}}
    \State Encoding Fact Structure:

    $\mathbf h_f^{(s)} = \textsc{Structural-Encoder}(\mathcal G, \phi, \psi; \mathbf a_P, \mathbf q_A)$
\EndIf

\State Attribute Score, Loss:

$\mathbf o_f^{(a)}, \mathcal L_f^{(a)} = \textsc{Scoring-Function}(\mathbf h_f^{(a)}; \{\mathbf h_s^{(a)} \mid s \in S\}; \mathbf w_S)$

\State Alignment Score, Loss:

$\mathbf o_f^{(l)}, \mathcal L_f^{(l)} = \textsc{Scoring-Function}(\mathbf h_f^{(a)}; \{\mathbf h_s^{(s)} \mid s \in S\}; \mathbf w_S)$

\If {training {\bf is True}}
    \State Structural Score, Loss:

    $\mathbf o_f^{(s)}, \mathcal L_f^{(s)} = \textsc{Scoring-Function}(\mathbf h_f^{(s)}; \{\mathbf h_s^{(s)} \mid s \in S\}; \mathbf w_S)$
\Else
    \State $\mathcal L_f^{(s)} = 0$
\EndIf

\State Final Loss:

$\mathcal{L}_f = \theta_a \mathcal{L}_f^{(a)} + \theta_s \mathcal{L}_f^{(s)} + \theta_l \mathcal{L}_f^{(l)}$

\State Final Predictions:

$\hat{\mathbf y_f} = I \left(\lambda_a \mathbf o_f^{(a)} + \lambda_l \mathbf o_f^{(l)} \geq \tau \right)$

\end{algorithmic}
\end{algorithm}

Algorithm~\ref{alg:main} details the forward pass of our proposed model \model{}, the same as was depicted in Figure~\ref{fig:arch}.
The attribute encoder converts texts to vector representations for both Facts and Sections.
The structural encoder is used to generate another set of vector representations for the Sections (at all times) and Facts (only train time).

The scoring function can use any pair of embeddings to generate scores and losses. It is used to calculate the attribute and alignment scores (at all times) and the structural scores (only train time).
These various scores and losses are combined to get the final outputs of the model.

\section{Citation network \& Metapaths} \label{sec:meta}


Here we take a brief look at the heterogeneous legal citation network and the metapaths used in \model{}.

\subsection{Overview of the Citation Network}

\begin{figure}[h]
\centering
\includegraphics[width=0.8\columnwidth]{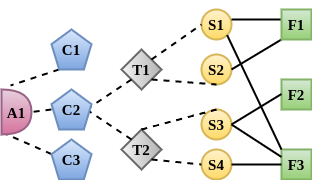} 
\caption{Zoomed-in illustration of the Citation Network in the architecture diagram of Figure~\ref{fig:arch}}
\label{fig:cnet}
\vspace{-6mm}
\end{figure}

The citation network can be decomposed into the two (overlapping) parts:

\noindent (1)~{\bf Label Hierarchy Tree:} As described in Section~\ref{sec:data} (example in Table~\ref{tab:ipchier}), the entire {\it Act}, IPC, can be divided into coarse-grained categories called {\it Chapters} and further subdivided into fine-grained categories called {\it Topics}. 
Each Topic is a non-overlapping grouping of {\it Sections} related to the same crime. This hierarchy of Acts, Chapters, Topics and then Sections form the Label Tree (see Figure~\ref{fig:cnet}). 
The nodes at each level have an {\it includes} relationship with the nodes below them in the hierarchy, and a {\it part of} relationship in the other direction.

\noindent (2)~{\bf Fact-side Graph:} The citations between Sections and Facts (documents) can be considered as a bi-partite graph.
In fact, just this part alone is sufficient to describe the citation network; we include the Label hierarchy as part of the entire network to exploit hierarchical relationships between the Sections themselves.
Each Fact has a {\it cites} relationship with Sections it is linked to, and there is a {\it cited by} relationship the other way.

\begin{table*}[!t]
\centering
\small
\begin{tabular}{|p{0.1\textwidth}|p{0.8\textwidth}|}
\hline
{\bf Facts of the case} & 

``[PERSON 1] was child of [PERSON 2] and his wife [PERSON 3].
[PERSON 2] and [PERSON 3] use to reside in the neighbourhood of [PERSON 4], original accused.

On the fateful day, i.e., on [DATE 1], [PERSON 3] was sitting alongwith her child in the courtyard of [PERSON 5]'s house, who is husband of original accused [PERSON 4].

Her husband [PERSON 2] came there and asked his wife to go with him to feed fertilizer to the standing crop in the field. 
However, [PERSON 3] declined to go with him as there was nobody to lookafter [PERSON 1]. 
Hearing this, [PERSON 4] and other accused persons told that they will look after the child and she could leave the child with them.
[PERSON 6] and one [PERSON 7] were also there at that time.
The child was left by [PERSON 2] with the accused persons and she left the place alongwith her husband. 

When both of them returned at about 4 O'clock, they straightaway went to the house of [PERSON 5] and enquired about [PERSON 1].
Accused told them that [PERSON 1] was playing in the vicinity only. 
However, they could not find the child. 
They searched for the child for the rest of the day and night and even on [DATE 2], but to no effect. 

However, in the morning of [DATE 3], when [PERSON 2] was searching for child alongwith [PERSON 5], they searched the house or hut of [PERSON 5] and saw that in the northern corner of that hut, foot of small child was protruding out of the ground. 
Seeing this [PERSON 1] reached to the Police Station, [PLACE 1] and reported the matter.'' \\

\hline 

{\bf Section 201} &

{\it Causing disappearance of evidence of offence, or giving false information to screen offender}

``Whoever, knowing or having reason to believe that an offence has been committed, causes any evidence of the commission of that offence to disappear, with the intention of screening the offender from legal punishment, or with that intention gives any information respecting the offence which he knows or believes to be false; 

shall, if the offence which he knows or believes to have been committed is punishable with death, be punished with imprisonment of either description for a term which may extend to seven years, and shall also be liable to fine;

and if the offence is punishable with imprisonment for life, or with imprisonment which may extend to ten years, shall be punished with imprisonment of either description for a term which may extend to three years, and shall also be liable to fine; 

and if the offence is punishable with imprisonment for any term not extending to ten years, shall be punished with imprisonment of the description provided for the offence, for a term which may extend to one-fourth part of the longest term of the imprisonment provided for the offence, or with fine, or with both.''
\\ 

\hline

{\bf Section 302} &

{\it Punishment for murder}

``Whoever commits murder shall be punished with death, or imprisonment for life, and shall also be liable to fine.'' \\

\hline

\end{tabular}%
\vspace{-3mm}
\caption{Example of the Fact Section taken from the Indian Supreme Court Document {\bf ``Liyakat vs State Of Uttaranchal on 25 February, 2008''}, along with the IPC Sections (201 and 302) that it cites.}   
\label{tab:example}
\end{table*}

\subsection{Description of Metapath Schemas}

As described in Section~\ref{sec:method-se}, a metapath is a sequence of nodes and the relationships connecting them in a heterogeneous network. 
A `metapath schema' is a user-defined sequence of node types and their connecting relations. 
Individual metapath instances can be constructed for a target node, and are sampled from the graph following the metapath schema.

Metapath schemas are defined keeping a target node in mind. All defined metapath schemas start and end with the same node type, and are defined in a way to extract meaningful relationships between the metapath neighbours.

In our work, we use 4 Fact-side metapaths and 4 Section-side metapaths, as described below.

\vspace{2mm}
\noindent {\bf Section-side metapaths:} Since $S$ type nodes are connected to both $T$ and $F$ type nodes, metapath schemas can be designed in both directions.

\noindent (1)~$S \xrightarrow{ctb} F \xrightarrow{ct} S$

\noindent (2)~$S \xrightarrow{po} T \xrightarrow{inc} S$

\noindent (3)~$S \xrightarrow{po} T \xrightarrow{po} C \xrightarrow{inc} T \xrightarrow{inc} S$

\noindent (2)~$S \xrightarrow{po} T \xrightarrow{po} C \xrightarrow{po} A \xrightarrow{inc} C \xrightarrow{inc} T \xrightarrow{inc} S$

\vspace{2mm}
\noindent {\bf Fact-side metapaths:} Since $F$ type nodes are connected to only $S$ type nodes, all defined metapath schemas must pass through nodes of type $S$.

\noindent (1)~$F \xrightarrow{ct} S \xrightarrow{ctb} F$

\noindent (2)~$F \xrightarrow{ct} S \xrightarrow{po} T \xrightarrow{inc} S \xrightarrow{ctb} F$

\noindent (3)~$F \xrightarrow{ct} S \xrightarrow{po} T \xrightarrow{po} C \xrightarrow{inc} T \xrightarrow{inc} S \xrightarrow{ctb} F$

\noindent (4)~$F \xrightarrow{ct} S \xrightarrow{po} T \xrightarrow{po} C \xrightarrow{po} A \xrightarrow{inc} C \xrightarrow{inc} T \xrightarrow{inc} S \xrightarrow{ctb} F$

\section{Example of Input \& Output} \label{sec:exp}

To illustrate the nature and complexity of the LSI problem, we show a sample Fact (input).
As described in Section~\ref{sec:method-ae}, we treat text in our problem as a nested sequence of sentences and words.

Table~\ref{tab:example} gives an illustration of the Facts of a case, and the relevant IPC Sections (gold-standard).
Table~\ref{tab:out} states the outputs of our model \model{} and of two closest baselines DEAL~\cite{hao2020inductive} and DPAM~\cite{wang2018modeling} on the input Fact shown in Table~\ref{tab:example}.
\model{} gets the output completely correct (identifies both the relevant Sections, and does not predict anything extra) while the other models fail to identify at least one relevant Section (and also identify wrong Sections marked in red).

\begin{table}[tb]
\centering
\small
\begin{tabular}{|l|p{0.6\columnwidth}|}
\hline
\textbf{Model} & \textbf{Predictions} \\ \hline
DEAL & \textcolor{red}{\bf Sec 173, Sec 190, Sec 200} \\
DPAM & \textcolor{red}{\bf Sec 200}, \textcolor{green}{\bf Sec 302} \\
LeSICiN & \textcolor{green}{\bf Sec 201, Sec 302} \\ \hline
\end{tabular}%
\vspace{-3mm}
\caption{Comparing outputs of {\bf LeSICiN} and of the two closest baselines {\bf DEAL} and {\bf DPAM}, on the input Fact shown in Table~\ref{tab:example}. 
Sections marked in \textcolor{green}{\bf green} are correctly identified (according to the gold-standard), and those in \textcolor{red}{\bf red} are wrongly identified.}   
\label{tab:out}
\end{table}

Although DPAM cannot get the entire result correct, it  correctly identifies one Section. But DEAL performs poorly on this example -- it could not identify any of the relevant Sections; also the Sections predicted by DEAL are \textit{not} very similar to the gold-standard Sections 201 and 302. 

\end{appendices}
\end{document}